%% file: lod-2025.tex
\documentclass[runningheads]{llncs}

\usepackage[T1]{fontenc}
\usepackage{amsmath}
\usepackage{amssymb}
\usepackage{amsfonts}
\usepackage{algorithm}
\usepackage{algpseudocode}
\usepackage{graphicx}
\usepackage{booktabs}
\usepackage{multirow}
\usepackage{siunitx}
\usepackage{array}
\usepackage{geometry}

\begin{document}
\title{Explorations of the Softmax Space: Knowing When the Neural Network Doesn't Know}
%
%
\author{Daniel Sikar\orcidID{0009-0001-3070-9542} \and
Artur d'Avila Garcez\orcidID{0000-0001-7375-9518} 
\and Tillman Weyde\orcidID{0000-0001-8028-9905}}
\authorrunning{D. Sikar et al.}
%
\institute{Department of Computer Science, City St George's, University of London, UK \\
\email{daniel.sikar,a.garcez,t.e.weyde@citystgeorges.ac.uk}
}
\maketitle              
%
\input{01-Abstract}



\input{02-Introduction}

\input{03-Context}

\input{04-Methods}
\input{05-ResultsAndDiscussion}

\input{06-ConclusionsAndFutureWork}
\bibliographystyle{splncs04}
\bibliography{references.bib}
%






\end{document}

%% file: 01-Abstract.tex


\begin{abstract}
Ensuring the reliability of automated decision-making based on neural networks will be crucial as Artificial Intelligence systems are deployed more widely in critical situations. This paper proposes a new approach for measuring confidence in the predictions of any neural network that relies on the predictions of a softmax layer. We identify that a high-accuracy trained network may have certain outputs for which there should be low confidence. In such cases, decisions should be deferred and it is more appropriate for the network to provide a \textit{not known} answer to a corresponding classification task. Our approach clusters the vectors in the softmax layer to measure distances between cluster centroids and network outputs. We show that a cluster with centroid calculated simply as the mean softmax output for all correct predictions can serve as a suitable proxy in the evaluation of confidence. Defining a distance threshold for a class as the smallest distance from an incorrect prediction to the given class centroid offers a simple approach to adding \textit{not known} answers to any network classification falling outside of the threshold. We evaluate the approach on the MNIST and CIFAR-10 datasets using a Convolutional Neural Network and a Vision Transformer, respectively.
The results show that our approach is consistent across datasets and network models, and indicate that the proposed distance metric can offer an efficient way of determining when automated predictions are acceptable and when they should be deferred to human operators.  
\keywords{Uncertainty \and Softmax Probability Distribution Space \and Distribution Shift \and Machine Learning}
\end{abstract}

%% file: 02-Introduction.tex

\section{Introduction}

Ensuring the reliability and safety of automated decision-making systems is important, particularly in high-impact areas where there exists potential for significant harm from errors \cite{amodei2016concrete}. 
Machine learning (ML) models, while powerful, are susceptible to making erroneous predictions when faced with data that differs from the distribution that they were trained on \cite{hendrycks2021many}. 
This phenomenon, known as distribution shift, poses a significant challenge in deploying ML in real-world scenarios. Distribution shift is a pervasive issue in ML, occurring when the distribution of the data used to train a model differs from the distribution of the data that the model encounters during deployment \cite{quinonero2009dataset}. This discrepancy can lead to a significant degradation in model performance, as it may struggle to generalize to the new, unseen data. 
Distribution shifts can manifest in various forms such as covariate shift, concept drift, and domain shift. Covariate shift arises when the input data distribution changes while the conditional distribution of the output given the input remains the same \cite{shimodaira2000improving}. Concept drift occurs when the relationship between the input and output variables changes over time \cite{gama2014survey}. Domain shift refers to the situation where the model is trained on data from one domain but applied to data from a different domain \cite{patel2015visual}.

To quantify and address distribution shift, researchers have developed various metrics and techniques. One common approach is to use statistical divergence measures, such as Kullback-Leibler (KL) divergence \cite{kullback1951information} or Maximum Mean Discrepancy (MMD) \cite{gretton2012kernel}, to assess differences in training and test data distributions. These metrics provide a quantitative understanding of the extent of the distribution shift.
Another approach is to employ domain adaptation techniques, which aim to align the feature distributions of the source and target domains \cite{wang2018deep}. This can be achieved through methods such as importance weighting \cite{sugiyama2007covariate}, feature transformation \cite{pan2009survey}, or adversarial learning \cite{ganin2016domain}. These techniques seek to mitigate the impact of distribution shift by making the model more robust to changes in the data distribution.

In this paper, we propose a simple and efficient approach for quantifying the confidence in the predictions made by a neural network. We evaluate the proposed metric in different distribution shift scenarios. Since practice has shown that solving the distribution shift problem may be highly domain dependent, we seek to simply identify as many as possible low-confidence predictions, rather than prescribe a specific domain adaptation technique, allowing the ML system to flag when, instead of prescribing a given classification result, the answer should be \textit{unknown} and decisions should be deferred to a human expert. 

Our method uses clustering of the softmax vector space to measure the distances between the outputs of a trained network and cluster centroids. By analyzing the distances in the softmax space, we propose a metric that can provide insight about the confidence that one may assign to model predictions. We adopt the most conservative possible threshold value at which model predictions are expected to be 100\% accurate, but we also analyze the impact of other possible threshold choices of potential practical value. 

We evaluate empirically the relationship between the distance to cluster centroids and the model's predictive accuracy. Our findings confirm that classes predicted more accurately by the model tend to have lower softmax distances to the respective centroids. This points to the value of using the changes observed in the softmax distribution given new data as a proxy for confidence in the model's predictions. The overall objective is to establish a closer link than thus far identified by the literature between a distance-based metric proposed for evaluating distribution shifts using unsupervised learning and the approaches that measure the drop in model accuracy based on supervised learning. Our empirical results indicate a useful connection between model performance and reliability, that is, model accuracy and how confident one can be in the predictions of a neural network based on the proposed distance metric. 



The main contributions of this paper are:
\begin{itemize}
\item A new lightweight approach for quantifying the reliability of neural network outputs requiring minimal computational overhead, using only the existing softmax outputs and a set of centroids and thresholds, without additional network modifications or training.
\item A combined analysis of the distance metric and model accuracy to tackle distribution shifts with experimental results showing consistency of the proposed approach across CNN and ViT architectures.
\end{itemize}

The remainder of the paper is organized as follows. Section 2 contains the required background. Section 3 introduces the clustering and softmax distance method. Section 4 contains the experimental results and Section 5 concludes the paper and discusses directions for future work.

%% file: 03-Context.tex
\section{Background}




\textbf{Softmax prediction probabilities}: The idea of using the entire set of softmax prediction probabilities, rather than solely relying on the maximum output, has been extensively studied in the context of enhancing the safety, robustness, and trustworthiness of machine learning models. Considering the complete distribution of class predictions provided by the softmax output enables the exploration of uncertainty quantification and other anomaly detection techniques that go beyond point estimates \cite{gal2016dropout}.

Uncertainty quantification is a crucial aspect of reliable ML such as Bayesian networks, as it allows for the estimation of confidence in the model's predictions \cite{kendall2017uncertainties}. 
Such approaches help identify instances where the model is less confident, enabling the system to defer to human judgment or take a more conservative action in safety-critical scenarios \cite{michelmore2018evaluating}. Such approaches are also well-known for being computationally costly, which probably explains the current preference for less reliable but efficient neural networks with a softmax layer, applicable to very large data sets. 

Moreover, the softmax probabilities can be utilized for anomaly detection, which is essential for identifying out-of-distribution (OOD) samples or novel classes that the model has not encountered during training \cite{hendrycks17baseline}. By monitoring the softmax probabilities, thresholding techniques can be applied to detect anomalies based on the distribution of the predictions \cite{liang2018enhancing}. 

The use of softmax probabilities also facilitates the development of more robust models that can handle adversarial examples and other types of input perturbations \cite{goodfellow2014explaining}. Adversarial attacks aim to fool the model by crafting input samples that lead to incorrect predictions with high confidence \cite{szegedy2013intriguing}. By considering the entire softmax distribution, defensive techniques such as adversarial training \cite{madry2017towards} and input transformations \cite{guo2018countering} can be applied to improve the model's robustness against these attacks.


The importance of leveraging the entire softmax distribution extends to various domains, including autonomous vehicles \cite{michelmore2018evaluating}, medical diagnosis \cite{leibig2017leveraging}, and financial risk assessment \cite{feng2018deep}. In safety-critical applications, the consequences of incorrect predictions can be severe, and relying solely on the Maximum a Posteriori (MAP) selection rule may not provide sufficient safeguards to reduce the risk of catastrophic failures.

However, the use of softmax probabilities is not without challenges. The calibration of the model's predictions is an important consideration, as poorly calibrated models may lead to overconfident or underconfident estimates. Techniques such as temperature scaling \cite{guo2017calibration} and isotonic regression \cite{zadrozny2002transforming} have been used to improve the calibration of the softmax probabilities.


\noindent \textbf{Clustering}: Clustering algorithms are essential for discovering structures and patterns in data across various domains \cite{xu2015comprehensive}. K-means, a widely used clustering algorithm, efficiently assigns data points to the nearest centroid and updates centroids iteratively \cite{lloyd1982least}. However, it requires specifying the number of clusters and it can be sensitive to initial centroid placement \cite{arthur2007k}. Hierarchical clustering creates a tree-like structure by merging or dividing clusters \cite{johnson1967hierarchical} but may not scale well to larger datasets \cite{mullner2011modern}. Density-based algorithms \cite{ester1996density,schubert2017dbscan} identify clusters as dense regions separated by lower density areas. 

Clustering has had many applications including in image segmentation for object detection \cite{shi2000normalized}, anomaly detection for fraud and intrusion detection \cite{chandola2009anomaly}, customer segmentation and bioinformatics. The choice of algorithm depends on data characteristics, desired cluster properties, and computational resources. Given the standing of K-means as almost the \textit{standard approach} to clustering, we simply chose to use K-means in this paper. Despite its many forms and  applications, clustering does not seem to have been applied up until now as an unsupervised learning method in the study of the softmax vector space of trained neural networks.

%% file: 04-Methods.tex

\section{Softmax Clustering for Uncertainty Quantification}
\label{methods:clustering}

\sloppy

We consider a neural network output vector $\mathbf{p} = (p_1, p_2, \dots, p_K)$ where $\sum p_i = 1$, representing a probability distribution obtained by normalizing the logit vector $\mathbf{z} = (z_1, z_2, \dots, z_K)$ through the softmax function, $p_i = \text{softmax}(z_i) = e^{z_i} / \sum_{j=1}^{K} e^{z_j}$. For example, given $\mathbf{p} = [0.01, 0.01, 0.01, 0.01, 0.9, 0.01, 0.01, 0.01, 0.01, 0.01]$, the predicted class is '4', corresponding to the highest value at index five, reflecting the confidence of the prediction for each class from '0' to '9'. The logits, representing log-likelihoods of class memberships, are related to probabilities by $z_i = \log (p_i / (1 - p_i))$ where $z_i$ is the logit for class $i$, and $p_i$ is the probability of the input belonging to class $i$.

We store the predictions for MNIST and CIFAR-10 datasets in a matrix $\mathbf{M} \in \mathbb{R}^{n \times 12}$, where $n$ is the number of predictions, the first ten columns are the softmax probabilities, column 11 is the true class and column 12 is the predicted class.
To obtain cluster centroids $\mathbf{C} \in \mathbb{R}^{10 \times 10}$ we calculate the mean of all correct predictions from the training datasets with Algorithm \ref{alg:k-means-centroid-init}. To calculate the softmax distance threshold we use all incorrect predictions with Algorithm \ref{alg:min_distance}.

\begin{algorithm}
\caption{K-Means Centroid Initialisation from Softmax Outputs}
\label{alg:k-means-centroid-init} 
\begin{algorithmic}[1]
\Require{$correct\_preds$: array of shape $(n, 12)$, where $n$ is the number of correct predictions}
\Ensure{$centroids$: array of shape $(10, 10)$, initialised centroids for each digit class}

\State $probs\_dist \gets corrects\_preds[:, :10]$ \Comment{Extract probability distribution for each digit}
\State $centroids \gets \text{zeros}((10, 10))$ \Comment{Initialise centroids array}

\For{$digit \gets 0$ to $9$}
\State $indices \gets \text{where}(\text{argmax}(probs\_dist, \text{axis}=1) == digit)[0]$ \Comment{Find indices of rows where digit has highest probability}
\State $centroid \gets \text{mean}(probs\_dist[indices], \text{axis}=0)$ \Comment{Compute mean probability distribution for selected rows}
\State $centroids[digit] \gets centroid$ \Comment{Assign centroid to corresponding row in centroids array}
\EndFor

\State \textbf{return} $centroids$
\end{algorithmic}
\end{algorithm}


\textbf{Geometric Characteristics of Probability-Constrained Cluster Spaces}: Cluster centroids produced by  softmax layer outputs adhere to valid probability distributions. This requires each coordinate \( c_i \) to satisfy \( 0 \leq c_i \leq 1 \) and the sum \( \sum_{i=1}^n c_i = 1 \), confining the centroids to the vertices and interior of the standard \((n-1)\)-dimensional simplex. For two such centroids \( \mathbf{c}_1 \) and \( \mathbf{c}_2 \) within the \( n \)-dimensional unit hypercube under these constraints, the Euclidean distance between centroids \( \mathbf{c}_1, \mathbf{c}_2 \in [0,1]^n \) is \( d(\mathbf{c}_1, \mathbf{c}_2) = \sqrt{\sum_{i=1}^{n}\left(c_{1i} - c_{2i}\right)^2} \). When enforcing probabilistic constraints via softmax normalization, the maximum achievable distance between centroids remains bounded by \(\sqrt{2}\), independent of the dimensionality \( n \).

\begin{algorithm}
\caption{Find Minimum Softmax Distances to Centroids for Incorrectly Predicted Digits (Threshold)}
\label{alg:min_distance} 
\begin{algorithmic}[1]
\Procedure{FindMinDistances}{$data$}
    \State $labels \gets [0, 1, 2, 3, 4, 5, 6, 7, 8, 9]$
    \State $thresh \gets \text{empty array of shape } (10, 2)$
    \For{$i \gets 0 \text{ to } 9$}
        \State $label \gets labels[i]$
        \State $min\_dist \gets \min(data[data[:, 1] == label, 0])$
        \State $thresh[i, 0] \gets min\_dist$
        \State $thresh[i, 1] \gets label$
    \EndFor
    \State \textbf{return} $thresh$
\EndProcedure
\end{algorithmic}
\end{algorithm}


\textbf{Cluster Density}: Quantifying cluster density in high-dimensional softmax spaces provides insight into the confidence of neural network predictions. The density of a cluster in an $n$-dimensional space is quantified using the following equations:
\begin{equation}
\rho_k = \frac{N_k}{V_n(r_k)}, \quad V_n(r_k) = \frac{\pi^{n/2}}{\Gamma\left(\frac{n}{2} + 1\right)} r_k^n, \quad N_k = \sum_{i=1}^n \mathbf{1}(\|\mathbf{x}_i - \mathbf{c}\| \leq r_k) \label{eq:density_volume_count}
\end{equation}

\noindent where \(\rho_k\) is the cluster density for radius scaling factor \(k\) \cite{Ester1996}, \(N_k\) is the number of points within radius \(r_k\), \(V_n(r_k)\) is the volume of the \(n\)-dimensional hypersphere with radius \(r_k\) \cite{Conway1998}, \(r_k = k \cdot r\) is the scaled radius, \(k\) is the scaling factor, \(r\) is the base radius, \(n\) is the number of dimensions, \(\pi\) is the mathematical constant, \(\Gamma\) is the gamma function, \(\mathbf{1}\) is the indicator function, \(\mathbf{x}_i\) is the \(i\)-th point’s coordinates, \(\mathbf{c}\) is the cluster centroid, and \(\|\cdot\|\) is the Euclidean norm, measuring the distance between points \cite{Duda2000}. These enable evaluation of cluster compactness at varying radii \cite{Hastie2009}.



The volume $V_{shell}$ of a shell between inner and outer radii $r_1, r_2$  is given by:

\begin{equation}
V_{\text{shell}}(n, r_1, r_2) = \frac{\pi^{n/2}}{\Gamma(n/2 + 1)} \cdot \left( r_2^n - r_1^n \right)
\end{equation}

The number of points $N_{shell}$ in a shell is given by $ N_{shell}(r_1,r_2) = N_{r_2} - N_{r_1} $, where $N_{r_2}$ is the number of points within a hypersphere of radius $r_2$ and $N_{r_1}$ is the number of points within hypersphere of radius $r_1$, where both hyperspheres are concentric and $r_2 > r_1$.










In n-dimensional spaces, hypersphere shells demonstrate volume distribution patterns that differ significantly from low-dimensional intuition. For n > 5, the volume concentrates predominantly in outer shells, while exponentially decreasing toward the center. The ratio between consecutive shell volumes increases with dimension, creating a steep volume gradient. This concentration effect stems from the $r^n$ term in the above volume formula. This affects nearest-neighbor calculations and creates challenges for clustering algorithms with most points existing near the hypersphere's surface rather than distributed throughout its volume. To handle this issue, in our approach thresholds will be calculated w.r.t. the true and predicted classes in the training set.



\textbf{Network Models}: A Convolutional Neural Network (CNN) was used to classify handwritten digits from the MNIST dataset, consisting of 60,000 training images and 10,000 testing images, each of size 28x28 grayscale (single channel) pixels, representing digits from 0 to 9.\footnote{The CNN architecture, implemented using PyTorch, consists of two convolutional layers followed by two fully connected layers. The first convolutional layer has 16 filters with a kernel size of 3x3 and a padding of 1. The second convolutional layer has 32 filters with the same kernel size and padding. Each convolutional layer is followed by a ReLU activation function and a max-pooling layer with a pool size of 2x2. The output of the second convolutional layer is flattened and passed through two fully connected layers with 128 and 10 neurons, respectively. The final output is passed through a log-softmax function to obtain the predicted class probabilities.} The CNN was trained using the Stochastic Gradient Descent (SGD) optimizer with a learning rate of 0.01 and a batch size of 64. The learning rate was determined using a custom learning rate function that decreases the learning rate over time. The Cross-Entropy Loss function was used as the criterion for optimization. The model was trained for 10 epochs.

The MNIST dataset was preprocessed using a transformation pipeline that converted the images to PyTorch tensors and normalized the pixel values to have a mean of 0.5 and a standard deviation of 0.5. The dataset was then loaded using PyTorch's DataLoader, for batch processing and shuffling of the data. The total number of parameters for the MNIST classification CNN is 206,922 and it took 5m6s to train on a Dell Precision Tower 5810 with a 6-core Intel Xeon Processor and 32GB memory running Ubuntu 18.04.  

A Vision Transformer (ViT) architecture \cite{dosovitskiy2020image} was implemented using the Hugging Face Transformers \cite{wolf2020huggingfaces}
library.\footnote{The model used 'google/vit-base-patch16-224-in21k' \cite{wu2020visual}, is pre-trained on the ImageNet-21k dataset \cite{deng2009imagenet}, which contains 14 million labeled images. The pre-trained model is then fine-tuned on the CIFAR-10 dataset.
The ViT model divides an input image into patches and processes them using a Transformer encoder. The ViT used has a patch size of 16x16 and an image size of 224x224. It outputs a representation of the image, which is then passed through a linear layer to obtain the final class probabilities.} The ViT model has 86.4M parameters and is fine-tuned to classify images from the CIFAR-10 dataset, consisting of 50,000 training images and 10,000 testing images, each of size 32x32 pixels, representing 10 different classes: airplane, automobile, bird, cat, deer, dog, frog, horse, ship, and truck. The model is trained using the Hugging Face Trainer API. The learning rate is set to $2 \times 10^{-5}$, the per-device train batch size is 10, the per-device \textit{eval} batch size is 4, and the number of training epochs is 3. The weight decay is set to 0.01. The model is trained on Google Colab Pro, T4 GPU hardware accelerator with high RAM.

The CIFAR-10 data preprocessing pipeline involved on-the-fly data augmentation using the torchvision library \textit{transforms} module. The training data undergoes random resized cropping, random horizontal flipping, conversion to a tensor, and normalization. The validation and testing data are resized, center-cropped, converted to a tensor, and normalized. 

%% file: 05-ResultsAndDiscussion.tex

\section{Experimental Results and Discussion}

We now present and discuss the results for CIFAR-10/ViT and MNIST/CNN. Both dataset/architecture pairs show similar trends in our analysis of the results. For context, in the case of MNIST with 60,000 well-balanced training examples (each digit class has approximately 6,000 examples), we find 54 occurrences of 8's misclassified as 9's. We note that 0's are the least likely and 8's are the most likely digits to be misclassified. Other digits that are less likely to be misclassified are 1's, 6's and 5's. Digit 8 is likely to be misclassified as a 6 or 9, while digit 3 is likely to be misclassified as a 2, 5 or 9. One can see how, already by itself, this kind of analysis can offer some insight onto the behavior of specific outputs in relation to other outputs and our confidence in the results, differently from the overall headline classification accuracy figure that in the case of MNIST has been close to $99\%$ accuracy. 


\textbf{Clustering}: We observe that calculating the average softmax outputs for all the correct classifications provides a good initialization for the K-means algorithm centroids. Clustering converges quickly and assigns all but two of the approximately 59,000 correctly classified examples (in the case of MNIST/CNN) to the correct class centroids. For example, all digits 0 are assigned to the same cluster that we shall label \textit{cluster 0}. In this case, we say that the unsupervised clustering task presents \emph{high fidelity} to the supervised classification task, and in this case it acts as a good proxy for assessing confidence in its predictions.

We compute all pairwise distances between the 10-class centroids in the softmax space.\footnote{($\binom{10}{2}=45$ pairs; although this may be a computationally costly task with an increasing number of classes, we note that typically the number of classes is much smaller than the number of parameters and the data set.} The distances exhibit $d_{min}=1.368$, $d_{max}=1.402$, $\mu=1.389$ and $\sigma=0.009$. These values approach $\sqrt{2}$, the maximum distance between points in the probability-constrained simplex, indicating that the model distributes classes at near-maximal separation in the output space. The supervised learning task achieved this separation through gradient descent on the cross-entropy loss; the unsupervised learning task confirms what has been induced by the classification task given the high-fidelity of the results.

\begin{figure}[ht]
    \centering
    \includegraphics[width=0.99\columnwidth]{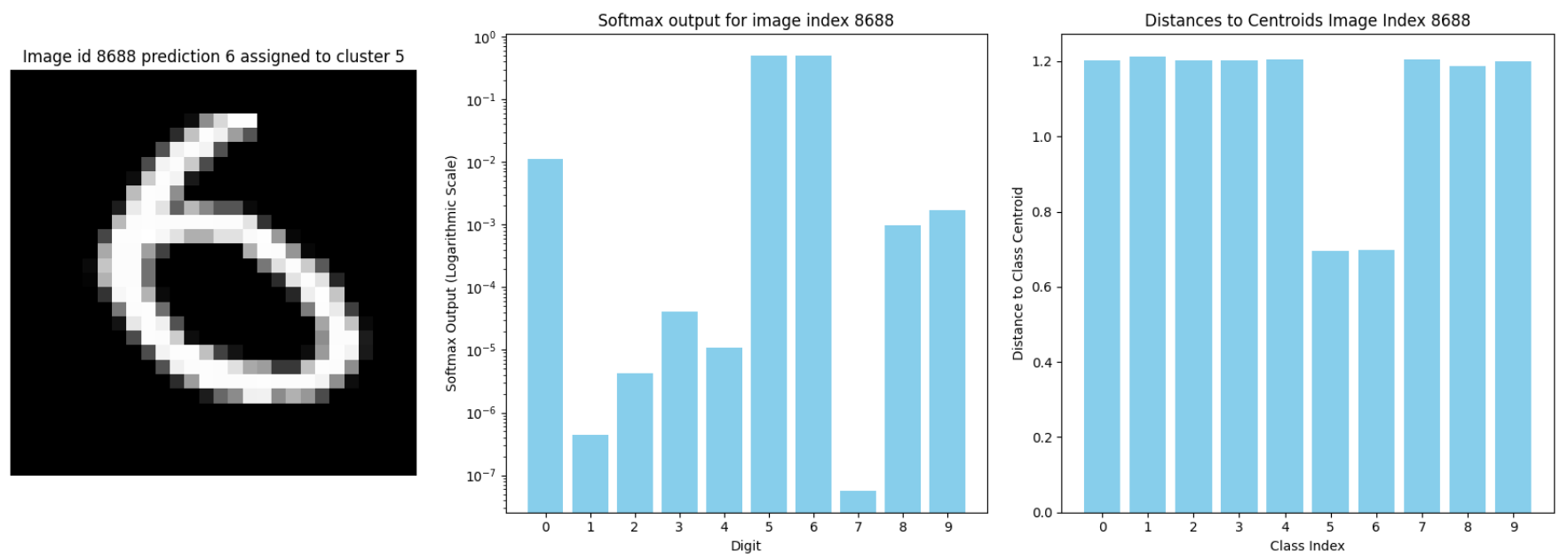}
    \caption{Left to right, MNIST Training Data Image ID 8688 digit 6, the network softmax output and the distances to class centroids, correctly classified as 6 and incorrectly clustered as 5. Notice that the y axis is not on logarithmic scale in this case.     
    }
\label{fig:ImageID8688_Softmax_Clustering}
\end{figure}

The above combination of supervised and unsupervised learning allows us to examine edge cases, where images are correctly classified but incorrectly clustered. The goal is to understand the softmax output better in connection with the distances to class centroids. Figure \ref{fig:ImageID8688_Softmax_Clustering} shows a correctly classified 6 that is incorrectly clustered as a 5, where the softmax outputs for digits 5 and 6 are very close (0.491 and 0.495, respectively), and the distances of that example to the centroid of clusters 5 and 6 are also very close (0.696 and 0.697, respectively). 
The softmax output shows high entropy, while the distances to class centroids show an example that is about 1.2 units away from all class centroids and close to 0.7 from both centroids 5 and 6. This illustrates a case where the model exhibits uncertainty, as would be expected from the softmax histogram already, but there are clear differences to the clustering histogram, despite the high-fidelity, to be investigated.


\begin{figure}[ht!]
    \centering
\includegraphics[width=0.99\columnwidth]{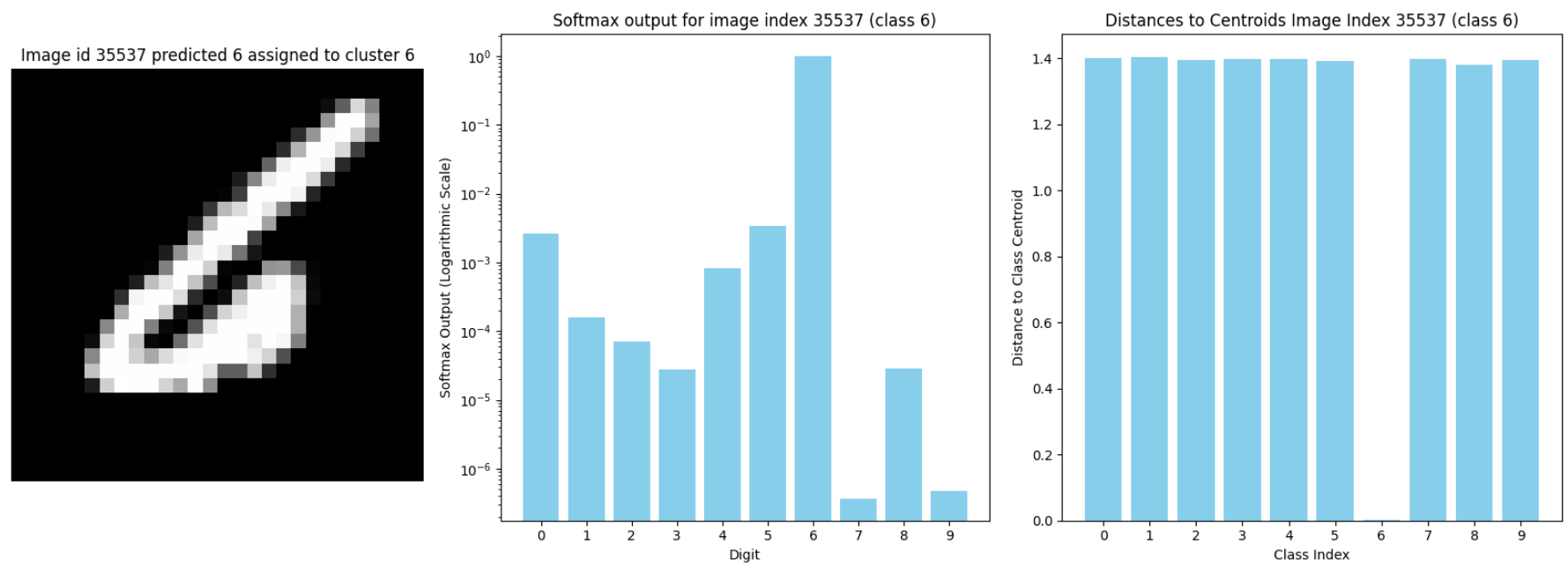}
    \caption{Left to right, MNIST Training Data Image ID 35537 digit 6, the network softmax output and the distances to class centroids, correctly classified and correctly clustered as 6.     
    }
\label{fig:ImageID35537_Softmax_Clustering}
\end{figure}

The next example presents a contrasting case to the previous edge case. Figure \ref{fig:ImageID35537_Softmax_Clustering} shows a 6 where the softmax output assigns 0.993 probability to class 6, with all other classes receiving probabilities below 0.004 (the y axis is in log scale). The low entropy of the softmax output matches a clear separation of centroid distances. As before, there are differences in the histograms to be further investigated. 

In what follows, more than investigating individual examples case-by-case as in Figures \ref{fig:ImageID8688_Softmax_Clustering} and \ref{fig:ImageID35537_Softmax_Clustering}, we will use the clustering (and hence the differences in the histograms) to seek to identify general trends for entire classes represented by the cluster centroids.

\textbf{Varying Thresholds}: The uncertainty that the above edge case exemplifies suggests that it would be reasonable to classify the data point, not as a 6 or a 5, but as \textit{unknown}. In fact, it could be argued that in this case the classification should be ``5 or 6", in that the image is likely to be of one of those two digits, but it is not known which. We now investigate, by varying the threshold used to define cluster membership, how certain examples may fall into an \textit{unknown} region of the clustering. We evaluate how accuracy varies by changing this threshold and also how many examples fall into the \textit{unknown} region (retention), as well as the ratios of correct to incorrect predictions. 

\begin{figure}[ht!]
    \centering
    \includegraphics[width=0.99\columnwidth]{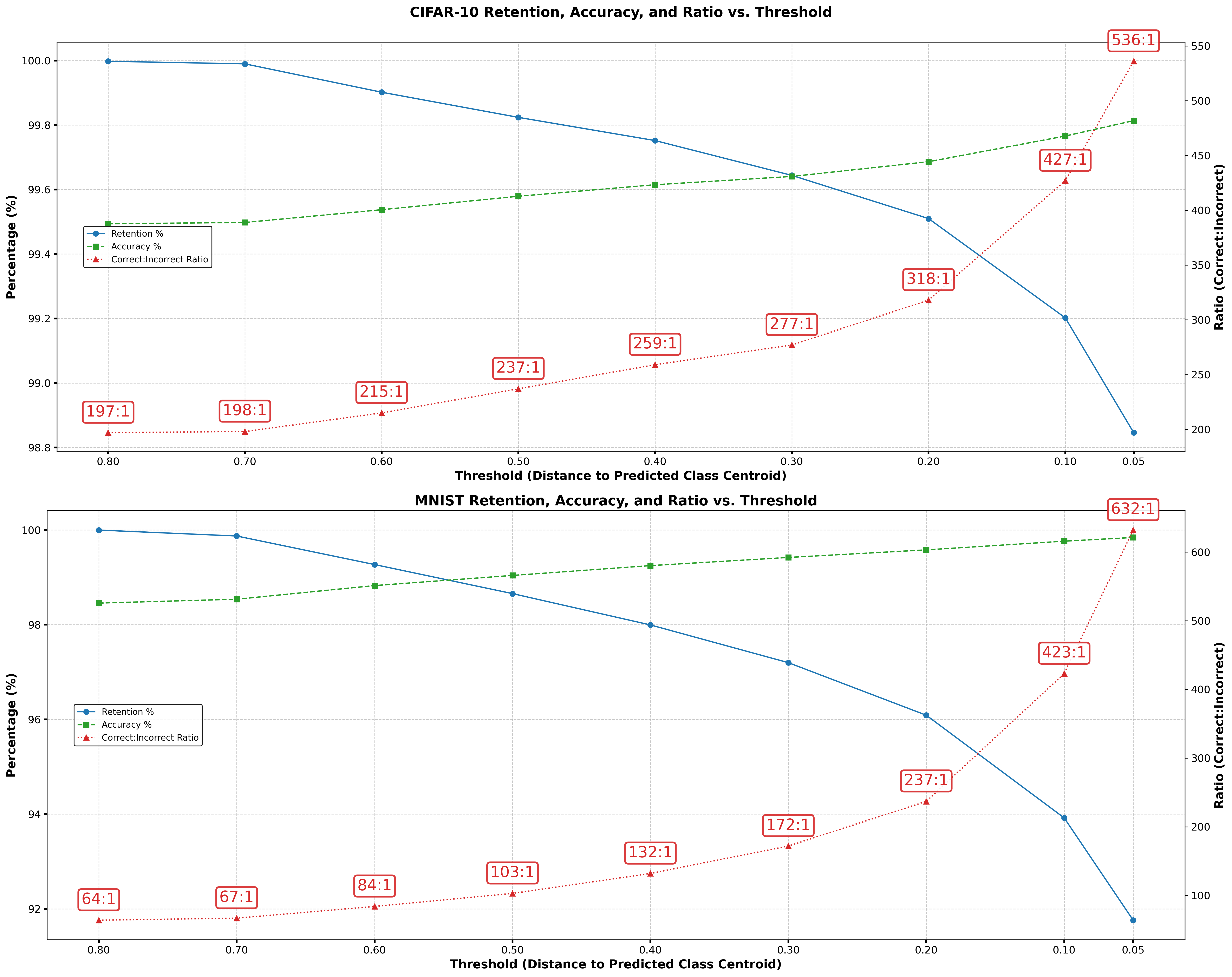}
    \caption{Retention, accuracy and correct-incorrect ratio vs Threshold ViT trained on CIFAR-10 and CNN trained on MNIST where the training dataset results are shown. The red plot represents the ratio of correct to incorrect predictions across all classes at given thresholds e.g. for CNN/MNIST at threshold 0.8 the ratio is 64:1, at threshold 0.05 the ratio is 632:1. The green plot is accuracy at every threshold e.g. at threshold 0.8 the accuracy is approximately 98.5\%, 64/(64+1), and at threshold 0.05 the accuracy is approximately 100\%, 632/(632+1). The blue plot represents the percentage of correct predictions that are being discarded as the threshold decreases e.g. at threshold 0.8 all correct predictions are kept while at threshold 0.05 over 8\% of the correct predictions are discarded i.e. classed as \textit{unknown}.}
\label{fig:MNIST_TRAINING_DATASET_ACC_RATIO_VS_THRESHOLD}
\end{figure}

Figure~\ref{fig:MNIST_TRAINING_DATASET_ACC_RATIO_VS_THRESHOLD} depict test set accuracy, retention (the percentage of examples belonging to the cluster hypersphere) and correct-to-incorrect ratio versus threshold (distance to predicted class centroid) for CIFAR/ViT (top) and MNIST/CNN (bottom). The y-left axis (percentage) ranges from 98.0\% to 100.0\% for both. For CIFAR/ViT, retention (blue) decreases from 100.0\% to 98.0\%, accuracy (green) increases from 99.4\% to 99.7\%, and the ratio (red) rises from 197:1 to 536:1 as the threshold decreases from 0.80 to 0.05. For MNIST/CNN, retention decreases from 100.0\% to 92.0\%, accuracy increases from 98.0\% to 99.0\%, and the ratio rises from 64:1 to 632:1 over the same threshold range. The ViT created more compact clusters, noting the ViT has 2 orders of magnitude more parameters than the CNN.


\begin{figure}[ht!]
    \centering
    \includegraphics[width=0.99\columnwidth]{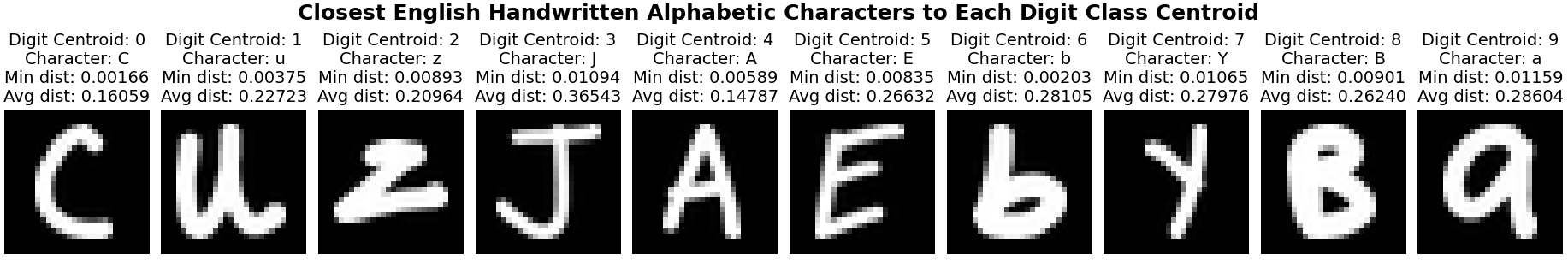}
    \caption{English Handwritten Alphabetic Characters nearest distance and example, and averages}
\label{fig:closest_alphabetic_characters_to_each_digit}
\end{figure}

\textbf{Out-of-distribution Data:} Predictions for in-distribution data are expected to fall closer to class centroids, and predictions for out-of-distribution data further away. To study out-of-distribution data predictions we \textit{MNISTify} the English Handwritten Characters dataset \cite{deCampos09} (separating digits from alphabetic characters for the sake of analysis), and the CIFAR-10 dataset. 
Figure \ref{fig:closest_alphabetic_characters_to_each_digit} shows which characters are nearest to each MNIST/CNN digit class centroid and also the average distance from the given class centroid (observing upper and lower case). For example, the lower case \textit{z} character shown (third from left) is among all alphabetic characters the nearest to centroid of digit class 2 (0.00893), where the average lower case \textit{z} is 0.20964 distance units away. Characters lower case \textit{b}, upper case \textit{B} and lower case \textit{a} show strong resemblance to digits \textit{6}, \textit{8} and \textit{9}, respectively. This is reflected is the clustering of the alphabetic dataset. Notice that no further training or fine-tuning of the CNN took place here. We only used MNIST/CNN for inference given the new character data set. We are interested in identifying, in the case of out-of-distribution data, how reasonable a prediction may be on average (not individual cases), as in the case of lower case \textit{z} above, and which predictions should fall in the \textit{unknown} region. 

Table~\ref{tab:threshold_percentages} shows dataset percentages that fall bellow or at/above threshold with varying thresholds. For example, for threshold 0.8, all datasets except \textit{MNISTified} CIFAR-10 have 100\% of the examples below threshold. At the lowest threshold of 0.05, CIFAR-10 has the highest percentage of examples below threshold, indicating the tightest clustering, followed by MNIST, where the number of in-cluster examples starts dropping significantly at 0.3 threshold. The \textit{MNISTified} English Handwritten Digits are more tightly clustered than the English Alphabetical Characters, as the former bears a closer resemblance to the MNIST dataset, and the latter have no equivalent labels in the MNIST dataset and consequently all predictions are technically incorrect. Notice that for the \textit{MNISTified} CIFAR-10, another case of all predictions being incorrect by default, and bearing no resemblance whatsoever to digits, has 97\% of examples rejected (excluded from all the clusters) at a threshold of 0.05. 

\begin{table}[ht]
\centering
\caption{Percentage of examples below and at or above thresholds for distance to predicted class centroid across five datasets. Total examples: CIFAR-10 (50,000), MNIST (60,000), MNISTified English Handwritten Digits (550), MNISTified English Handwritten Alphabetical Characters (2,860), MNISTified CIFAR-10 (50,000).}
\label{tab:threshold_percentages}
\begin{tabular*}{\textwidth}{@{\extracolsep{\fill}} c *{10}{S[table-format=3.1]}}
\toprule
 & \multicolumn{2}{c}{CIFAR-10} & \multicolumn{2}{c}{MNIST} & \multicolumn{2}{c}{Eng. Digits} & \multicolumn{2}{c}{Eng. Alphabetical} & \multicolumn{2}{c}{MNISTified CIFAR-10} \\
\cmidrule(lr){2-3} \cmidrule(lr){4-5} \cmidrule(lr){6-7} \cmidrule(lr){8-9} \cmidrule(lr){10-11}
Threshold & {Below} & {At/Above} & {Below} & {At/Above} & {Below} & {At/Above} & {Below} & {At/Above} & {Below} & {At/Above} \\
\midrule
0.8  & 100.0 & 0.0  & 100.0 & 0.0  & 100.0 & 0.0  & 100.0 & 0.0  & 98.2  & 1.8  \\
0.7  & 100.0 & 0.0  & 99.9  & 0.1  & 99.3  & 0.7  & 97.4  & 2.6  & 85.5  & 14.5 \\
0.6  & 99.9  & 0.1  & 99.3  & 0.7  & 96.5  & 3.5  & 90.2  & 9.8  & 67.4  & 32.6 \\
0.5  & 99.8  & 0.2  & 98.7  & 1.3  & 93.1  & 6.9  & 83.0  & 17.0 & 52.6  & 47.4 \\
0.4  & 99.8  & 0.2  & 98.0  & 2.0  & 89.3  & 10.7 & 75.0  & 25.0 & 39.1  & 60.9 \\
0.3  & 99.6  & 0.4  & 97.2  & 2.8  & 86.0  & 14.0 & 67.2  & 32.8 & 27.2  & 72.8 \\
0.2  & 99.5  & 0.5  & 96.1  & 3.9  & 81.6  & 18.4 & 58.0  & 42.0 & 16.5  & 83.5 \\
0.1  & 99.2  & 0.8  & 93.9  & 6.1  & 72.4  & 27.6 & 47.0  & 53.0 & 6.9   & 93.1 \\
0.05 & 98.8  & 1.2  & 91.8  & 8.2  & 67.8  & 32.2 & 38.7  & 61.3 & 3.0   & 97.0 \\
\bottomrule
\end{tabular*}
\end{table}

\begin{figure}[ht!]
    \centering
    \includegraphics[width=0.99\columnwidth]{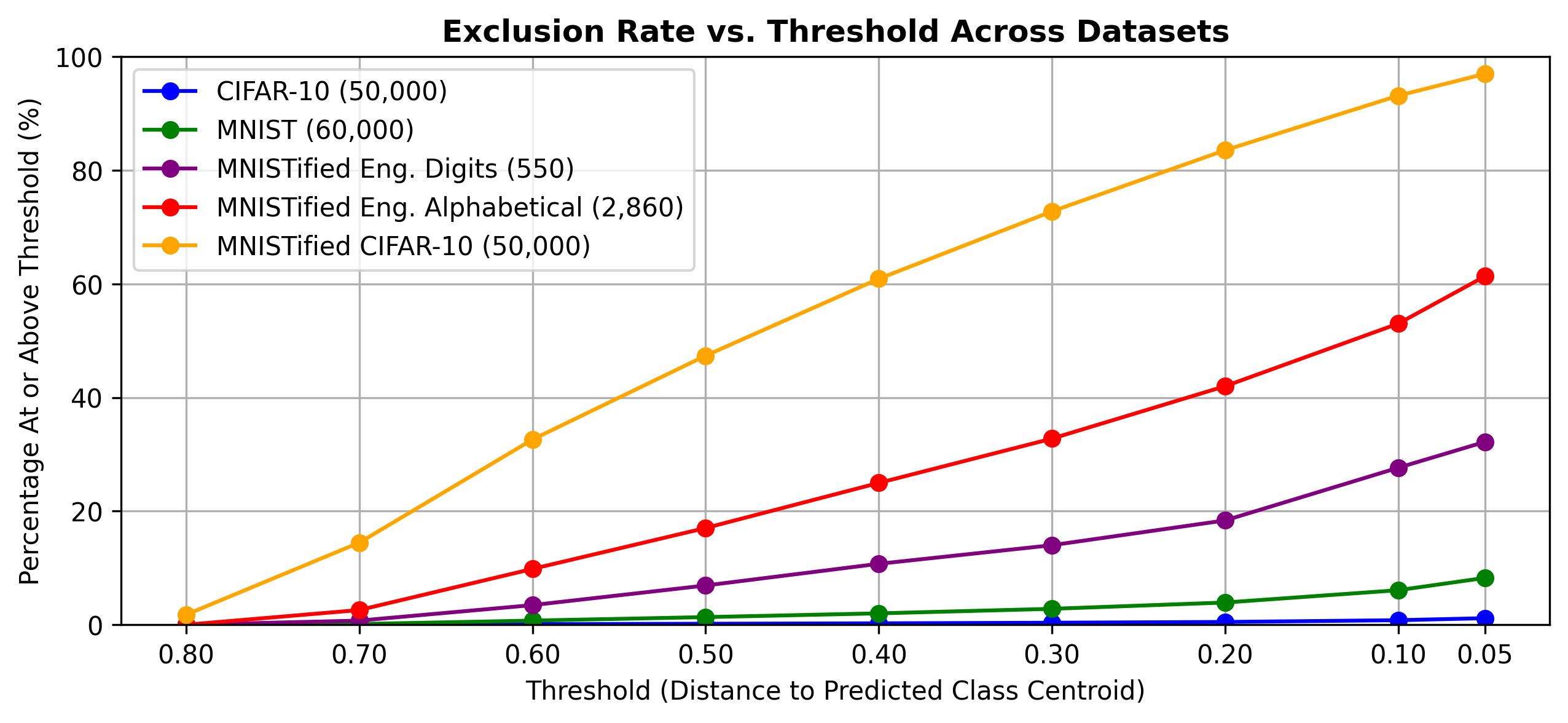}
    \caption{Percentage of examples at or above thresholds for distance to predicted class centroid across five datasets, showing higher exclusion rates for English Alphabetical Characters and MNISTified CIFAR-10. Total examples: CIFAR-10 (50,000), MNIST (60,000), Eng. Digits (550), Eng. Alphabetical (2,860), MNISTified CIFAR-10 (50,000).}
\label{fig:Exclusion_Rate_vs_Threshold}
\end{figure}

Figure~\ref{fig:Exclusion_Rate_vs_Threshold} shows the relationship between distance thresholds (x-axis) and exclusion percentages (y-axis). Five datasets are represented: CIFAR-10 (blue), MNIST (green), MNISTified English Digits (purple), MNISTified English Alphabetical (red), and MNISTified CIFAR-10 (yellow), with dataset counts indicated in parentheses. The x-axis ranges from 0.80 to 0.05, representing decreasing threshold values for the distance to predicted class centroid. The y-axis shows percentage values from 0\% to 100\%. MNISTified CIFAR-10 exhibits the highest exclusion rates across all thresholds. The MNISTified English Alphabetical dataset shows moderate exclusion rates (up to 61\%), while MNISTified English Digits demonstrates lower exclusion rates (up to 32\%). Both MNIST and CIFAR-10 maintain minimal exclusion rates below 10\% across all thresholds, with CIFAR-10 showing the lowest values overall, remaining under 1\%. 

\begin{table}[ht]
\centering
\small
\caption{Density of points (\(N_k\), \(\rho_k\)) in 10-dimensional spherical shells and inner sphere, based on distance to predicted class centroid. Higher densities in outer shells for English Alphabetical Characters and MNISTified CIFAR-10 indicates greater exclusion. Total examples: CIFAR-10 (50,000), MNIST (60,000), Eng. Digits (550), Eng. Alphabetical (2,860), MNISTified CIFAR-10 (50,000).}
\label{tab:density}
\begin{tabular}{@{} l l S[table-format=1.5e-1] *{5}{cc} @{}}
\toprule
 & & & \multicolumn{2}{c}{CIFAR-10} & \multicolumn{2}{c}{MNIST} & \multicolumn{2}{c}{Eng. Digits} & \multicolumn{2}{c}{Eng. Alphabetical} & \multicolumn{2}{c}{M'fied CIFAR-10} \\
\cmidrule(lr){4-5} \cmidrule(lr){6-7} \cmidrule(lr){8-9} \cmidrule(lr){10-11} \cmidrule(lr){12-13}
{Shell} & {Radii} & {Volume} & {$N_k$} & {$\rho_k$} & {$N_k$} & {$\rho_k$} & {$N_k$} & {$\rho_k$} & {$N_k$} & {$\rho_k$} & {$N_k$} & {$\rho_k$} \\
\midrule
1 & 0.7--0.8 & 2.01786e-1 & 4 & 2.0e1 & 75 & 3.7e2 & 4 & 2.0e1 & 74 & 3.7e2 & 6336 & 3.1e4 \\
2 & 0.6--0.7 & 5.6616e-2 & 44 & 7.8e2 & 362 & 6.4e3 & 15 & 2.6e2 & 207 & 3.7e3 & 9077 & 1.6e5 \\
3 & 0.5--0.6 & 1.29295e-2 & 39 & 3.0e3 & 368 & 2.8e4 & 19 & 1.5e3 & 206 & 1.6e4 & 7379 & 5.7e5 \\
4 & 0.4--0.5 & 2.22299e-3 & 36 & 1.6e4 & 397 & 1.8e5 & 21 & 9.4e3 & 227 & 1.0e5 & 6767 & 3.0e6 \\
5 & 0.3--0.4 & 2.52346e-4 & 54 & 2.1e5 & 478 & 1.9e6 & 18 & 7.1e4 & 223 & 8.8e5 & 5934 & 2.4e7 \\
6 & 0.2--0.3 & 1.47973e-5 & 67 & 4.5e6 & 666 & 4.5e7 & 24 & 1.6e6 & 264 & 1.8e7 & 5383 & 3.6e8 \\
7 & 0.1--0.2 & 2.60882e-7 & 154 & 5.9e8 & 1301 & 5.0e9 & 51 & 2.0e8 & 315 & 1.2e9 & 4802 & 1.8e10 \\
8 & 0.05--0.1 & 2.54767e-10 & 178 & 7.0e11 & 1296 & 5.1e12 & 25 & 9.8e10 & 238 & 9.3e11 & 1919 & 7.5e12 \\
Inner & $\leq$0.05 & 2.49039e-13 & 49423 & 2.0e17 & 55056 & 2.2e17 & 373 & 1.5e15 & 1106 & 4.4e15 & 1513 & 6.1e15 \\
\bottomrule
\end{tabular}
\end{table}

\begin{figure}[ht]
    \centering
\includegraphics[width=0.99\columnwidth]{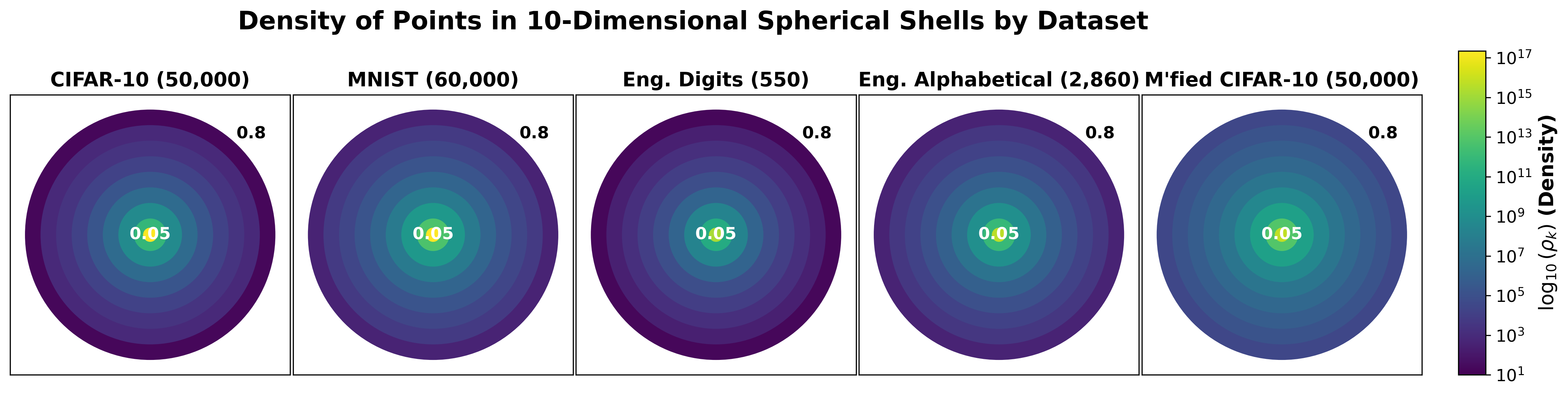}
\caption{Density of points in 10-dimensional spherical shells, visualized as concentric rings. Color intensity represents $\log_{10}(\rho_k)$, with brighter colors indicating higher density. English Alphabetical Characters and MNISTified CIFAR-10 show higher density in outer shells, reflecting greater exclusion. Total examples: CIFAR-10 (50,000), MNIST (60,000), Eng. Digits (550), Eng. Alphabetical (2,860), MNISTified CIFAR-10 (50,000).}
\label{fig:density_rings}
\end{figure}

\textbf{Cluster Density}: Figure~\ref{fig:density_rings} presents a concentric ring visualization of the density of points (\(\rho_k\)) in 10-dimensional spherical shells, defined by distance thresholds to the predicted class centroid [0.7--0.8, 0.6--0.7, 0.5--0.6, 0.4--0.5, 0.3--0.4, 0.2--0.3, 0.1--0.2, 0.05--0.1] and an inner sphere (\(\leq 0.05\)), across five datasets: CIFAR-10 (50,000 examples), MNIST (60,000 examples), English Handwritten Digits (550 examples), English Handwritten Alphabetical Characters (2,860 examples), and MNISTified CIFAR-10 (50,000 examples). The figure, a row of 5 subplots with a shared colorbar, encodes the density of softmax outputs in the 10-dimensional probability simplex, highlighting the higher exclusion rates (larger distances) for English Alphabetical Characters and MNISTified CIFAR-10. Next we examine the figure’s design, interpret its density patterns, and address the implications and limitations of projecting high-dimensional data into a 2D representation: each subplot represents a dataset, with eight concentric annuli corresponding to the shells and a central circle for the inner sphere. The radii are scaled to the thresholds (0.8 to 0.05), ensuring the outermost shell (0.7--0.8) is wider than the innermost (0.05--0.1), reflecting the hyperspherical geometry where volume scales as \(V_k \propto r^{10}\). Density (\(\rho_k\)) is mapped to a logarithmic color scale, \(\log_{10}(\rho_k)\), using the \texttt{viridis} colormap, with dark purple for low density (\(\sim 10^1\)) and bright yellow for high density (\(\sim 10^{17}\)). This logarithmic scale accommodates the extreme density range, driven by the rapid volume decrease from \(2.02 \times 10^{-1}\) (Shell 1) to \(2.49 \times 10^{-13}\) (inner sphere). Annotations label the outermost ring (0.8) and inner sphere (0.05), and dataset names with example counts provide context. The shared colorbar, ranging from \(10^1\) to \(10^{17}\), ensures consistent density interpretation across subplots. The figure emphasizing higher exclusion for English Alphabetical Characters and MNISTified CIFAR-10. For CIFAR-10 and MNIST, the outer rings are dark purple, indicating low density (\(\rho_1 = 2.0 \times 10^1\), \(\rho_2 = 7.8 \times 10^2\) for CIFAR-10; \(\rho_1 = 3.7 \times 10^2\), \(\rho_2 = 6.4 \times 10^3\) for MNIST), with few points at larger distances (\(N_1 = 4\), \(N_2 = 44\) for CIFAR-10; \(N_1 = 75\), \(N_2 = 362\) for MNIST). Their inner spheres are intensely yellow (\(\rho_9 = 2.0 \times 10^{17}\), \(N_9 = 49423\) for CIFAR-10; \(\rho_9 = 2.2 \times 10^{17}\), \(N_9 = 55056\) for MNIST), reflecting tight clustering near class centroids, consistent with low exclusion rates (1.2\% and 8.2\% at or above 0.05). The fine-tuned Vision Transformer (ViT) for CIFAR-10 and the convolutional neural network (CNN) optimized for MNIST produce confident predictions, concentrating softmax outputs in the inner sphere.

Conversely, English Alphabetical Characters and MNISTified CIFAR-10 display brighter outer rings, signaling higher density at larger distances. For English Alphabetical Characters, densities values like \(\rho_1 = 3.7 \times 10^2\), \(\rho_2 = 3.7 \times 10^3\), and \(\rho_4 = 1.0 \times 10^5\) appear as purple to green rings, driven by significant point counts (\(N_1 = 74\), \(N_2 = 207\), \(N_4 = 227\)). This reflects the MNIST CNN’s uncertainty in mapping letters to digit centroids, resulting in a high exclusion rate (61.3\% at or above 0.05). MNISTified CIFAR-10 shows even brighter green to yellow outer rings (\(\rho_1 = 3.1 \times 10^4\), \(\rho_2 = 1.6 \times 10^5\), \(\rho_4 = 3.0 \times 10^6\); \(N_1 = 6336\), \(N_2 = 9077\), \(N_4 = 6767\)), corresponding to the highest exclusion rate (97.0\% at or above 0.05), as the CNN misaligns non-digit CIFAR-10 images. English Handwritten Digits exhibit moderate outer-ring brightness (\(\rho_1 = 2.0 \times 10^1\), \(\rho_2 = 2.6 \times 10^2\); \(N_1 = 4\), \(N_2 = 15\)) and a less intense inner sphere (\(\rho_9 = 1.5 \times 10^{15}\), \(N_9 = 373\)), aligning with an intermediate exclusion rate (32.2\%).

Figure~\ref{fig:density_rings} underscores the relationship between model performance and prediction geometry. The ViT’s robustness for CIFAR-10 and the CNN’s optimization for MNIST yield dense inner spheres, indicating high confidence and low exclusion. The CNN is tested for out-of-distribution data (letters, non-digit images) and produces denser outer shells, suggesting potential for distance-based anomaly detection. However, the 2D projection simplifies the 10-dimensional probability simplex, where softmax outputs sum to 1, and the hyperspherical model assumes Euclidean distances, potentially skewing absolute density values. The extreme volume disparity (\(10^{-1}\) to \(10^{-13}\)) amplifies inner-sphere density, and the logarithmic colormap may compress subtle outer-shell differences. In high dimensions, the ``curse of dimensionality'' concentrates volume near hypersphere surfaces, implying that even in-distribution points may lie in thin shells, though small distances (\(\leq 0.05\)) place them in the inner sphere.

%% file: 06-ConclusionsAndFutureWork.tex
\section{Conclusion and Future Work}

This paper introduced a computationally lightweight approach for quantifying the confidence and reliability of predictions obtained from any neural network utilizing a softmax output layer. By clustering the softmax probability vectors and measuring the Euclidean distance between a given output and class centroids derived from the mean of correct predictions, we established a practical metric for assessing prediction confidence. A key finding is that this distance metric effectively serves as a proxy for confidence, enabling the system to identify potentially unreliable predictions and defer judgment by providing an ``unknown" answer, even when the original network has not been trained with that additional class, when a prediction falls outside a defined distance threshold. This threshold can be conservatively set based on the minimum distance observed for incorrect predictions within the training data. In practice, the choice of threshold will be domain dependent, although the analysis provided in Figure \ref{fig:Exclusion_Rate_vs_Threshold} may help an expert decide where the threshold should be set, based on the safety requirements or constraints of the application. The analysis provided in Figure \ref{fig:MNIST_TRAINING_DATASET_ACC_RATIO_VS_THRESHOLD} may also provide subsidies for the choice of threshold in the case of out-of-distribution data when expert knowledge is available comparing different data sets (or in the case where the data can be separated as with English digits and characters).

Our empirical evaluations, conducted using both Convolutional Neural Networks on MNIST and Vision Transformers on CIFAR-10, demonstrated consistency of results across different network architectures and datasets and the effectiveness of the proposed approach. The results confirmed that accurately predicted, in-distribution examples tend to cluster more tightly around their respective class centroids compared to misclassified or out-of-distribution examples, as shown in the analyses involving MNISTified datasets. The analysis of cluster density in the high-dimensional softmax space further reinforced results, showing higher densities in outer shells (larger distances) for OOD data and supporting the earlier results indicating robustness of the ViT model. We expect this method to offer a valuable tool in the analysis of trustworthiness of ML systems, flagging low-confidence predictions with minimal computational overhead. The approach is being evaluated on self-driving scenarios using the CARLA simulator.